\documentclass[10pt,twocolumn,letterpaper]{article}

\usepackage[pagenumbers]{cvpr}
\usepackage[ruled, vlined]{algorithm2e}
\usepackage{array}
\usepackage{multirow}
\usepackage{pgfplots}
\usepackage{xcolor}
\usepackage{caption}

\newcommand{\UBA}{\operatorname{UBAttn}}
\newcommand{\RMSNorm}{N_{\text{RMS}}}
\newcommand{\SwiGLU}{\operatorname{SwiGLU}}

\newlength{\rowlabw}

\newlength{\imgw}

\definecolor{cvprblue}{rgb}{0.21,0.49,0.74}

\definecolor{plotblue}{RGB}{47,125,190}
\definecolor{plotred}{RGB}{220,53,69} 
\pgfplotsset{compat=1.18}
\usetikzlibrary{shapes.geometric,arrows.meta,calc}
\pgfdeclareplotmark{fstar}{\node[star, star points=5, star point ratio=2.4, inner sep=0pt, minimum size=2.6\pgfplotmarksize, fill] {};}

\usepackage[pagebackref,breaklinks,colorlinks,allcolors=cvprblue]{hyperref}

\title{Hierarchical Reasoning Action Head for Tiny Vision-Language-Action Models}

\author{Aditya Makineni \qquad Amitesh Sahoo \qquad Qing Tian\thanks{Corresponding author.}\\
Department of Computer Science, University of Alabama at Birmingham\\
{\tt\small \{amakinen, sahooa, qtian\}@uab.edu}
}

\begin{document}
\maketitle
\begin{abstract}
Real-time onboard robot control calls for a lightweight vision-language-action (VLA) model, yet shrinking the backbone leaves less capacity for deliberation at inference time. Chain-of-thought decoding, dual-system architectures, and larger backbones can restore deliberation but reintroduce the latency and memory costs that motivated going small. Recurrent action heads offer a lighter alternative by iterating in latent space before committing to an action, but existing designs refine a single undifferentiated state that must simultaneously encode slowly evolving task context and rapidly changing motor commands. We propose the Hierarchical Reasoning Action Head (HR-AH), which maintains two latent states updated at different frequencies: a low-level state that updates fast, grounded in action tokens and proprioception, and a high-level state that updates slowly, grounded in visual understanding and the instruction-informed low-level state. The two states exchange information through nested reasoning cycles. Because all iteration occurs in the compact head after a single backbone pass, HR-AH adds hierarchical reasoning without increasing the backbone size. With a 22.5M-parameter head, 3--4.3$\times$ smaller than those of RD-VLA and VLA-Adapter on the same backbone, HR-AH reaches 97.0\% success rate on LIBERO and a 4.40 average chain length on CALVIN ABC$\rightarrow$D, at the lowest latency among compared methods. We show that this hierarchy improves performance over a parameter-matched single-state head, and that routing distinct modalities to each level yields further gains in both accuracy and inference cost.
\end{abstract}    
\section{Introduction}
\label{sec:intro}

Vision-language-action (VLA) models attach an action decoder to a pretrained vision-language backbone, bringing general-purpose robotic manipulation directly from raw images and natural-language commands within reach~\cite{kim2024openvla, black2025pi0, kim2025oft}. However, deploying these models onboard at control frequency demands a compact backbone. A growing line of tiny VLAs delivers this by shrinking the backbone to a fraction of the original scale~\cite{wen2024tinyvla, belkhale2024minivla, shukor2025smolvla, wang2026vlaadapter}, but a smaller backbone has less capacity available for deliberation at inference time. Chain-of-Thought decoding at 3--8.5B parameters~\cite{zhao2025cot, zhong2026flowvla, huang2025thinkact, huang2026fast} and dual-system designs that distribute modalities across large VLM subsystems~\cite{zhang2024hirt, chen2025fastinslow} can restore deliberation, but they reintroduce the very costs that a tiny backbone was chosen to avoid. Recurrent action heads offer a lighter path by iterating in latent space before committing to an action~\cite{tur2026rdvla, shen2026loopvla}. Latent iteration suits chunked continuous control well, since the objective is to refine a motor plan rather than to generate tokens sequentially. However, latent recurrence alone does not determine how computation should be organized. Within an action chunk, task-level context remains relatively steady while concrete motor commands must be jointly refined across timesteps. Refining them in a single recurrent state forces it to represent quantities that evolve at different rates, analogous to running a task planner at servo rate~\cite{tur2026rdvla, shen2026loopvla}. This raises a central question: can explicitly separating the two improve the effectiveness and efficiency of recurrent computation without increasing the backbone's computational burden?

\begin{figure*}[t]
    \centering
    \begin{tikzpicture}[
        img/.style={inner sep=0pt, anchor=south west},
        frm/.style={draw=black!35, line width=0.3pt},
        tag/.style={anchor=north west, xshift=0.7mm, yshift=-0.7mm, fill=white, fill opacity=0.85, text opacity=1, inner sep=1.5pt, rounded corners=1pt, font=\scriptsize},
        rowlab/.style={rotate=90, align=center, font=\footnotesize, anchor=center},
    ]
    \definecolor{objA}{RGB}{206,104,22}\definecolor{objB}{RGB}{37,99,170}
    \def\w{3.0}\def\g{0.1}\def\gx{0.75}\def\xz{4.0}
    \foreach \f/\t [count=\c from 0] in {rd-vla-fc-1/{Step~1}, rd-vla-fc-4/{Step~4}, rd-vla-fc-8/{Step~8 (halted)}} {
        \node[img] (a\c) at ({\xz+\c*(\w+\gx)}, {\w+\g}) {\includegraphics[width=\w cm, trim=41 41 41 41, clip]{Images/Attn_viz/\f.png}};
        \draw[frm] (a\c.south west) rectangle (a\c.north east);
        \node[tag] at (a\c.north west) {\t};
    }
    \foreach \f/\t [count=\c from 0] in {hrah-fc-1/{Step~1}, hrah-fc-2/{Step~2}, hrah-fc-3/{Step~3 (halted)}} {
        \node[img] (b\c) at ({\xz+\c*(\w+\gx)}, 0) {\includegraphics[width=\w cm, trim=41 41 41 41, clip]{Images/Attn_viz/\f.png}};
        \draw[frm] (b\c.south west) rectangle (b\c.north east);
        \node[tag] at (b\c.north west) {\t};
    }
    \foreach \c in {1,2} {\foreach \d in {-1,0,1} {\fill[black!80] ({\xz+\c*(\w+\gx)-\gx/2+\d*0.14}, {1.5*\w+\g}) circle (0.65pt);}}
    \foreach \c in {1,2} {\draw[-{Stealth[length=3pt,width=2.5pt]}, black!70, line width=0.5pt] ({\xz+\c*(\w+\gx)-\gx/2-0.2}, {0.5*\w}) -- ({\xz+\c*(\w+\gx)-\gx/2+0.2}, {0.5*\w});}
    \node[rowlab] at ({\xz-0.38}, {1.5*\w+\g}) {Single state\\reasoning (RD-VLA)};
    \node[rowlab] at ({\xz-0.38}, {0.5*\w}) {Our hierarchical\\reasoning (HR-AH)};
    \node[inner sep=0pt] (in) at ({\w/2}, {\w+\g/2+0.3}) {\includegraphics[width=\w cm, trim=41 41 41 41, clip]{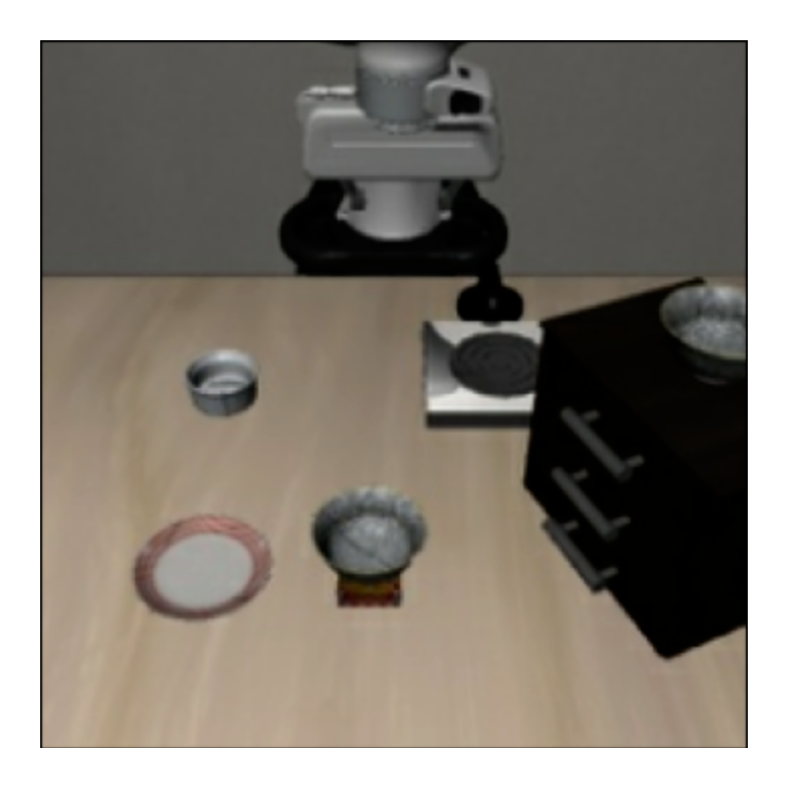}};
    \draw[frm] (in.south west) rectangle (in.north east);
    \def\px#1#2{($(in.south west)+({#1/706*\w},{(706-#2)/706*\w})$)}
    \foreach \col/\xa/\ya/\xb/\yb in {objA/330/330/330/440, objB/80/400/142/476} {
        \draw[-{Stealth[length=5pt,width=4.5pt]}, \col, line width=1.3pt] \px{\xa}{\ya} -- \px{\xb}{\yb};
    }
    \node[anchor=south, font=\footnotesize\bfseries, yshift=1pt] at (in.north) {Input observation};
    \pgfmathsetlengthmacro\tw{\w cm-6pt}
    \node[anchor=north, draw=black!40, rounded corners=2pt, fill=black!3, inner sep=3pt, text width=\tw, align=center, font=\footnotesize, yshift=-4pt] at (in.south) {\textbf{Task:} \textit{pick up the \textcolor{objA}{black bowl} on the cookie box and place it on the \textcolor{objB}{plate}}};
    \draw[-{Stealth[length=4pt,width=3.6pt]}, black!55, line width=0.6pt] (\xz, -0.3) -- ({\xz+3*\w+2*\gx}, -0.3) node[midway, below, font=\footnotesize, text=black!70, inner sep=2pt] {reasoning steps};
    \end{tikzpicture}
    \caption{Attention maps over reasoning steps for RD-VLA (top row) and our HR-AH (bottom row). The dashed ellipse highlights the task-relevant region. HR-AH halts after only 3 steps with attention sharply focused on the target bowl and destination plate, whereas the single-state baseline requires 8 steps and attends more diffusely. Both rows visualize attention from the vision-grounded latent state. Only the third-person view is shown.}
    \label{fig:qualitative_comparison}
\end{figure*}

We introduce the \textbf{Hierarchical Reasoning Action Head (HR-AH)}, which decouples the two processes into levels that correspond to naturally distinct inputs. The motor command draws on instruction-conditioned action queries and proprioception, while the task context draws on vision and the evolving instruction-informed motor plan. This arrangement, fast proprioception-driven feedback operating beneath slower vision-driven deliberation, reflects a well-established principle in sensorimotor control~\cite{todorov2002optimal, merel2019hierarchical} and in layered robot control architectures~\cite{brooks1986subsumption}. Unlike dual-system designs that distribute this hierarchy across two large VLM subsystems~\cite{zhang2024hirt, chen2025fastinslow}, HR-AH routes modalities to two states inside a single lightweight action head, keeping the entire model compact.

To our knowledge, HR-AH is the first hierarchical recurrent action head for a tiny VLA. The two levels exchange information through nested reasoning cycles with top-down refinement. Because vision tokens are the larger input, confining them to the slower path amortizes their cost over multiple fast executor steps. The backbone is evaluated once per observation, and all iterative reasoning takes place in the lightweight head. \cref{fig:qualitative_comparison} visualizes how the resulting attention sharpens over reasoning steps compared to a single-state baseline.
The contributions of this paper are as follows:
\begin{itemize}
    \item \textbf{Hierarchical recurrent computation for tiny VLAs.} HR-AH separates visual-context update from motor-command refinement through high- and low-level latent states, providing an explicit hierarchical structure for iterative action refinement entirely within the action head.
    \item \textbf{Modality-specific routing for efficient reasoning.} We route vision tokens to the high-level planner and action tokens with proprioception to the low-level executor, concentrating expensive visual attention in the slower reasoning pathway. Combined with adaptive halting, this routing allows inference depth to vary with task difficulty while keeping per-step cost low.
    \item \textbf{Strong results with a compact head.} With a 22.5M-parameter head, 3--4.3$\times$ smaller than the heads of RD-VLA and VLA-Adapter on the same backbone, HR-AH reaches 97.0\% on LIBERO, a 4.40 average chain length on CALVIN ABC$\rightarrow$D, and 78.8\% on a real Kinova Gen3 Lite robot, at the lowest latency among compared methods. Ablations show that hierarchical state separation accounts for most of the gain over a parameter-matched single-state baseline, while modality-specific routing contributes a further 3.4 points and reduces the number of reasoning steps required.
\end{itemize}

\begin{figure*}[t]
    \centering
    \includegraphics[width=\textwidth]{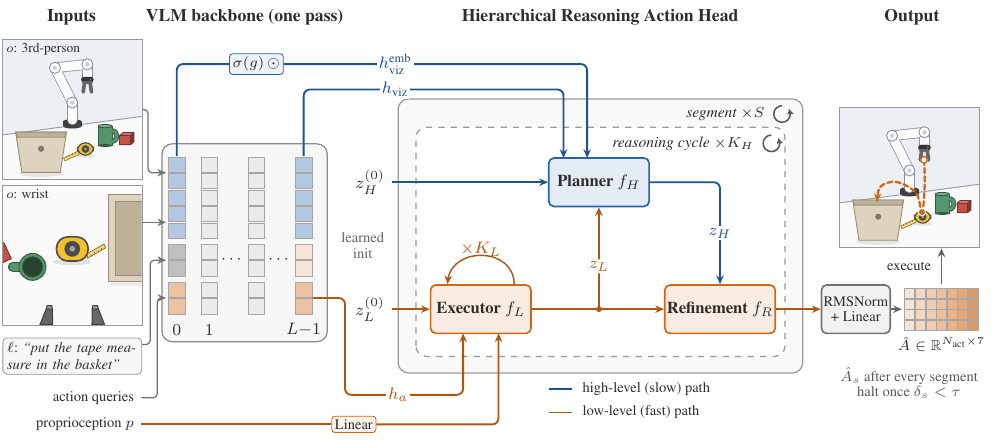}
    \caption{Overview of the Hierarchical Reasoning Action Head (HR-AH). The VLM backbone extracts multimodal features from a deep layer ($h_{\text{viz}}$, $h_a$) and the embedding layer ($h_{\text{viz}}^{\text{emb}}$). Inside the HR-AH, iterative latent reasoning is performed over $S$ segments, each comprising $K_H$ hierarchical reasoning cycles. Within each cycle, the executor ($f_L$) attends to concise action and proprioceptive context for frequent updates of the detailed motor commands, while the planner ($f_H$) processes the larger vision context and the instruction-informed compact $z_L$ at a slower rate to maintain an abstract task representation. The refinement module ($f_R$) feeds the updated plan back into the action state, closing the bidirectional loop. All three modules share the same $\UBA$ block architecture (\cref{sec:modules}) with different context inputs. After each segment $s$, the action state is projected to a candidate action chunk $\hat{A}_s$, and the reasoning halts once the relative change $\delta_s$ between consecutive predictions falls below a threshold $\tau$ (\cref{sec:inference}). The final prediction is executed as $\hat{A} \in \mathbb{R}^{N_{\text{act}} \times 7}$.
    }
    \label{fig:method}
\end{figure*}

\section{Related Work}
\label{sec:related}

\paragraph{Vision-language-action models.}
VLAs adapt pretrained vision-language models to robot control. They differ chiefly in how actions are decoded, e.g., whether by autoregressive discretization~\cite{kim2024openvla}, compressed action tokens~\cite{pertsch2025fast}, diffusion~\cite{chi2023diffusionpolicy}, flow matching~\cite{black2025pi0}, or parallel decoding with $\ell_1$ regression~\cite{kim2025oft}. A complementary line shrinks the backbone for practical deployment~\cite{wen2024tinyvla, belkhale2024minivla, shukor2025smolvla}. Most relevant is VLA-Adapter~\cite{wang2026vlaadapter}, which trains a feed-forward policy over a tiny backbone by injecting per-layer VLM features through Bridge Attention. Its 97M-parameter head is purely feed-forward, with no provision for iterative refinement. We operate in the same tiny-VLA setting but show that a compact hierarchical recurrent head can surpass feed-forward decoding in both accuracy and efficiency.

\paragraph{Iterative computation for action prediction.}
A growing body of work allocates variable computation per action. Early-exit methods adapt how much of the backbone runs at each step~\cite{yue2024deervla}, and dual-system designs split slow reasoning and fast control into distinct models or modules~\cite{shi2025hirobot, nvidia2025groot, bu2024robodual, zhang2024hirt}, with Fast-in-Slow~\cite{chen2025fastinslow} distributing modalities across two subsystems that share a 7B VLM at a 1:4 update ratio. UniFS~\cite{sun2026unifs} schedules backbone layer groups at decreasing update frequencies, LoopVLA~\cite{shen2026loopvla} repeatedly applies a shared transformer block to the multimodal tokens, and PredVLA~\cite{sawada2026predvla} is a sub-million-parameter multi-timescale recurrent policy with no VLM backbone. These approaches organize computation at different frequencies across backbone components or outside the action head. Recurrence inside the action head has been explored, but only with a single state. For example, RD-VLA~\cite{tur2026rdvla} places a single weight-tied recurrent core and shows that success scales with test-time iterations. Such a head uses the same recurrent representation to integrate visual context and refine motor commands at every iteration, rather than assigning these operations distinct update frequencies. No existing design combines multi-level hierarchy with iterative recurrence inside the action head. We close this gap with two latent states, updated at different frequencies after a single backbone pass, whose modality-routed inputs let scene context and motor commands be updated in separate but interacting streams. We show this design outperforms a parameter-matched single-state recurrent head with access to the same visual context, and competes with much larger architectures on standard benchmarks.

\paragraph{Recursive reasoning models.}
Adaptive and weight-tied recurrent computation was explored early by adaptive computation time~\cite{graves2016act} and Universal Transformers~\cite{dehghani2019universal}, and has resurfaced as recurrent-depth language models~\cite{geiping2025huginn} and looped transformers for 3D vision~\cite{burzio2026dejaview}. HRM~\cite{wang2025hrm} couples a slow high-level and a fast low-level recurrent module on symbolic puzzles. Its high-level module observes only the low-level module's output, so both levels are ultimately grounded in the same representation despite differing in update frequency. Subsequent analyses find that this hierarchy adds little on these tasks~\cite{arcprize2025hrm, ge2025hrmmisconceptions, rodrigues2026dissecting}, and TRM~\cite{jolicoeur2025trm} outperforms it with a single tiny network trained by backpropagating through the recursion. Embodied control hints at a different outcome, as PredVLA reports that removing the temporal hierarchy from its predictive recurrent architecture lowers success substantially~\cite{sawada2026predvla}. Our hierarchy routes distinct modalities to each level, so the timescale separation arises naturally from the distinct input modalities. We ablate against a parameter-matched single-state head that sees the full context from the backbone and a variant that preserves both states and their update rates but feeds them identical inputs. Hierarchy accounts for the larger gain, while modality-specific routing contributes a further improvement in both accuracy and inference cost.

\paragraph{Multi-layer VLM features for control.}
Which backbone depth best conditions an action head remains unsettled. GR00T~N1 conditions on a middle layer~\cite{nvidia2025groot}, SmolVLA discards the upper half of the backbone~\cite{shukor2025smolvla}, and FLOWER prunes layers and fuses intermediate tokens~\cite{reuss2025flower}. VLA-Adapter matches each policy block to one VLM layer~\cite{wang2026vlaadapter} and LIRA widens this to a local window of layers~\cite{zhang2026lira}, while ABot-M0 reports that, after large-scale robot pretraining, final-layer features alone suffice~\cite{yang2026abotm0}. Our head reads both token groups from a deep backbone layer and additionally supplies the planner with gated vision features from the embedding layer. This retains both high-level semantic features and fine-grained spatial detail while extracting all representations once per observation.

\section{Methodology}
\label{sec:method}

\subsection{Overview}
\label{sec:overview}
We consider a manipulation policy $\pi_{\theta}$ that maps visual observation $o$ (third-person and wrist camera views), language instruction $\ell$, and proprioceptive states $p$ to an action chunk $\hat{A} \in \mathbb{R}^{N_{\text{act}} \times d_a}$, where $N_{\text{act}}$ is the chunk length and $d_a$ is the action dimension. Unlike standard VLAs that map the final hidden state directly to actions, or recurrent heads that refine a single undifferentiated latent~\cite{tur2026rdvla}, we propose the \textbf{Hierarchical Reasoning Action Head (HR-AH)} as illustrated in \cref{fig:method}, which decouples the latent state into two interacting components updated at different frequencies:
\begin{itemize}
    \item A \textbf{low-level action state} $z_L \in \mathbb{R}^{N_{\text{act}} \times D}$, where $D$ is the hidden dimension, grounded in action tokens and proprioception, representing the evolving motor command.
    \item A \textbf{high-level planning state} $z_H \in \mathbb{R}^{N_{\text{plan}} \times D}$, where $N_{\text{plan}}$ is the number of planning tokens, grounded in vision tokens and the instruction-conditioned $z_L$, encoding task-relevant visual context to guide motor-command refinement.
\end{itemize}
This routing is deliberate. The action tokens provide a compact, instruction-conditioned input that, together with proprioception, directly grounds the motor command and feeds the low-level state at every step. Vision tokens are the larger input and change slowly, so routing them (together with $z_L$) to the high-level planner amortizes their cost over multiple low-level updates. The refinement step then lets vision guide the executor without requiring it to attend to vision tokens at every step.

Both states ($z_H$ and $z_L$) are initialized from learnable parameters $Z_H$ and $Z_L$ (truncated normal initialization), with sequence order maintained by Rotary Positional Embeddings (RoPE)~\cite{su2024roformer} within the attention layers. States are reinitialized at every chunk boundary. The overall pipeline extracts features from the backbone VLM once per observation (\cref{sec:features}), then iteratively refines $(z_H, z_L)$ through $S$ segments of $K_H$ hierarchical reasoning cycles each before projecting the refined $z_L$ to actions (\cref{eq:output}):
\begin{equation}
F = \operatorname{VLM}(o, \ell), \quad \hat{A} = \text{HR-AH}(F, p),
\label{eq:pipeline}
\end{equation}
where $F$ denotes the backbone features.

\subsection{Feature Extraction}
\label{sec:features}
We extract features from the VLM backbone to serve as input to the HR-AH. Given features $F$ from a chosen backbone stage (e.g., penultimate layer), we extract vision tokens $h_{\text{viz}}$ and action-query tokens $h_a$:
\begin{equation}
h_{\text{viz}} = F[:N_{\text{viz}}, :], \quad
h_a = F[-N_q:, :],
\label{eq:features}
\end{equation}
where $N_{\text{viz}}$ is the number of vision tokens and $N_q$ is the number of action-query tokens (typically $N_{\text{viz}} \gg N_q$). Language-token positions lie between the two groups and are not extracted directly, as is standard in causal VLAs where action queries already attend to the preceding language tokens~\cite{kim2025oft, wang2026vlaadapter, tur2026rdvla}. Proprioceptive input $p$ is projected into dimension $D$ via a linear layer.
The planner receives $h_{\text{viz}}$ as visual context. To supplement it with fine-grained spatial features that contextual layers may discard, we additionally extract vision tokens from the backbone embedding layer and apply a learned sigmoid gate:
\begin{equation}
h_{\text{viz}}^{\text{emb}} = \sigma(g) \odot F^{(0)}[:N_{\text{viz}}, :],
\label{eq:emb-gate}
\end{equation}
where $F^{(0)}$ denotes the features at the embedding layer (before any transformer block). $g \in \mathbb{R}^D$ is initialized so that $\sigma(g)$ is near zero, allowing the model to start from the deep features $h_{\text{viz}}$ alone and gradually learn how much embedding information to incorporate. All features are computed once per observation and reused across segments as context inputs to the reasoning modules (\cref{sec:modules}).

\subsection{Hierarchical Reasoning Modules}
\label{sec:modules}

\paragraph{Unified Bridge Attention Block.}
The Unified Bridge Attention ($\UBA$) block adapts the Bridge Attention mechanism~\cite{wang2026vlaadapter} to operate over arbitrary context sets. Given a query source $X \in \mathbb{R}^{n \times D}$ and a context set $\mathcal{C} = \{C_1, \dots, C_m\}$, $\UBA$ concatenates $X$ and all $C_i$ into one sequence and computes a single softmax so that $X$ attends jointly to itself and every context. Unlike Bridge Attention, $\UBA$ applies RoPE~\cite{su2024roformer} to queries and keys to encode token order within the concatenated sequence. $\UBA$ also removes the learnable gate on vision attention scores, making it source-agnostic, so $\mathcal{C}$ may be any set of sources and each source's contribution arises from query-key similarity alone. This source-agnostic design is what enables the hierarchical head: a single block definition can serve as the executor, planner, and refinement module, each receiving different contexts. Concretely, letting $\bar{C} = [X;\, C_1;\, \dots;\, C_m]$ denote the concatenated sequence,
\begin{equation}
\UBA(X,\,\mathcal{C}) = \operatorname{Attn}\!\bigl(X W_Q,\; \bar{C} W_K,\; \bar{C} W_V\bigr),
\label{eq:uba}
\end{equation}
where $\operatorname{Attn}$ is multi-head attention with projection matrices $W_Q, W_K, W_V$. Each block pairs $\UBA$ with a SwiGLU~\cite{shazeer2020glu} FFN under RMSNorm~\cite{zhang2019rmsnorm} ($\RMSNorm$) pre-normalization:
\begin{gather}
X' = X + \UBA\!\bigl(\RMSNorm(X),\, \{\RMSNorm(C_i)\}\bigr), \notag\\
\operatorname{Block}(X, \mathcal{C}) = X' + \SwiGLU(\RMSNorm(X')). \label{eq:block}
\end{gather}

\paragraph{Module routing.}
The HR-AH contains three modules, each implemented as a $\UBA$ block with different context inputs: $f_L$ (low-level executor), $f_H$ (high-level planner), and $f_R$ (top-down refinement). Concretely, the executor $f_L$ attends to the compact $h_a$ and $p$, enabling frequent updates ($K_L$ iterations per cycle) at low cost. The planner $f_H$ attends to the large $h_{\text{viz}}$, $h_{\text{viz}}^{\text{emb}}$, and the current $z'_L$, concentrating the costly vision attention at the slow rate while staying informed of the evolving motor command. The refinement module $f_R$ then injects the updated $z_H$ back into $z_L$, closing the bidirectional loop so that vision understanding reaches the motor command.

\subsubsection{Low-Level Executor}
\label{sec:low-level}
The low-level executor refines $z_L$ by attending to action tokens $h_a$ and proprioception $p$:
\begin{equation}
z'_L = f_L(z_L, \{h_a, p\}).
\label{eq:executor}
\end{equation}
Here and below, primes distinguish updated states within a single reasoning cycle. Each iteration grounds the action state in the action tokens $h_a$ and the robot's proprioceptive state $p$. Through $h_a$, the executor also embeds instruction information into $z'_L$, where language can be well represented due to its compact nature. 

\subsubsection{High-Level Planner}
\label{sec:high-level}
The high-level planner updates the planning state $z_H$ using the full vision tokens from different backbone depths and the instruction-informed action state $z'_L$:
\begin{equation}
z'_H = f_H(z_H, \{h_{\text{viz}},\, h_{\text{viz}}^{\text{emb}},\, z'_L\}).
\label{eq:planner}
\end{equation}
Concentrating the full vision context at the planner, which runs at the slow rate, avoids repeating the costly attention at every executor step. The planner integrates visual understanding into the planning state $z_H$, with $z'_L$ steering attention toward the features that matter for the current motor command.

\subsubsection{Top-Down Refinement}
\label{sec:refine}
Following the high-level update, we introduce a refinement step that explicitly injects the updated plan back into the low-level action state. The low-level state attends to itself and to the newly updated high-level state:
\begin{equation}
z''_L = f_R(z'_L, \{z'_H\}).
\label{eq:refine}
\end{equation}
This acts as a top-down modulation step, ensuring that the low-level motor command is aligned with the updated high-level intent before the next cycle or the final prediction.

\subsection{Segments and Hierarchical Reasoning Cycles}
\label{sec:segments}
The HR-AH performs iterative latent reasoning through a sequence of $S$ segments. Each segment consists of $K_H$ hierarchical reasoning cycles. A single cycle applies $K_L$ low-level executor iterations followed by one high-level planner iteration, and concludes with the top-down refinement. 
Starting from states $(z_H, z_L)$, one segment executes \cref{alg:segment}. To avoid buffering activations for all intermediate steps, we adopt the one-step gradient approximation from HRM~\cite{wang2025hrm}. Within each segment, only the last reasoning cycle is differentiable, with all preceding cycles running detached. At the end of each segment, we apply RMSNorm to both recurrent states to stabilize their magnitudes before carrying them into the next segment.

\begin{algorithm}[t]
    \DontPrintSemicolon
    \KwIn{States $(z_H, z_L)$; context $(h_{\text{viz}}, h_{\text{viz}}^{\text{emb}}, h_a, p)$; cycles $K_H$, iterations $K_L$}
    \KwOut{Updated states $(z_H, z_L)$}
    \BlankLine
    \For{$c = 1$ \KwTo $K_H$}{
        \For{$k = 1$ \KwTo $K_L$}{
            $z_L \leftarrow f_L(z_L, \{h_a, p\})$ \tcp*{Eq.~\ref{eq:executor}}
        }
        $z_H \leftarrow f_H(z_H, \{h_{\text{viz}}, h_{\text{viz}}^{\text{emb}}, z_L\})$ \tcp*{Eq.~\ref{eq:planner}}
        $z_L \leftarrow f_R(z_L, \{z_H\})$ \tcp*{Eq.~\ref{eq:refine}}
    }
    $z_L \leftarrow \RMSNorm(z_L)$\;
    $z_H \leftarrow \RMSNorm(z_H)$\;
    \caption{One segment of the HR-AH.}
    \label{alg:segment}
\end{algorithm}

After $S$ segments, the final low-level state is projected to the action chunk:
\begin{equation}
\hat{A} = \RMSNorm(z_L^{(S)})\, W_{\text{out}}.
\label{eq:output}
\end{equation}
During training, the model unrolls a fixed number of $S_{\text{total}}$ segments per forward pass. Each segment produces an action prediction $\hat{A}_s$, and we apply deep supervision with an $\ell_1$ loss between predicted and ground-truth action chunks, weighted linearly by segment index. States are detached between segments so that gradients flow only within each segment.

\subsection{Inference with Adaptive Halting}
\label{sec:inference}
Inspired by~\cite{tur2026rdvla}, we adopt a training-free halting criterion to dynamically determine the number of segments at inference time. After each segment $s$, the model computes the relative change in its action prediction:
\begin{equation}
\delta_s = \frac{\lVert \hat{A}_s - \hat{A}_{s-1} \rVert_\infty}{\lVert \hat{A}_s \rVert_\infty + \epsilon},
\end{equation}
where $\lVert \cdot \rVert_\infty$ denotes the maximum absolute value and $\epsilon$ is a small constant for numerical stability. Since no prior prediction exists at the first segment, the comparison begins at $s = 2$. Halting is triggered when $\delta_s < \tau$ for $P$ consecutive segments, subject to a minimum of $S_{\min}$ segments and a hard cap of $S_{\max}$. This allows the model to allocate more computation to complex observations while saving inference time on routine ones.

\section{Experiments and Results}
\label{sec:experiments}

\subsection{Experimental Setup}
\label{sec:setup}
We evaluate our method across standard simulation benchmarks and a real-world setting to assess both standardized performance and physical generalizability. For simulation, we train and evaluate on all four suites of the LIBERO benchmark~\cite{liu2023libero} (Spatial, Object, Goal, and Long) and the CALVIN ABC$\rightarrow$D benchmark~\cite{mees2022calvin}. For the real-world experiments, we collect a small dataset using a 6-DoF Kinova Gen3 Lite robot arm and evaluate the policy on a set of diverse manipulation tasks.

\paragraph{Implementation details.}
All models are trained on NVIDIA A100 GPUs. We use a MiniVLA backbone~\cite{belkhale2024minivla} built on Qwen2.5-0.5B~\cite{qwen2024qwen25} with fused DINOv2/SigLIP vision encoders, initialized from the MiniVLA checkpoint and fine-tuned with LoRA (rank 64). The action dimension $d_a = 7$, comprising end-effector translation deltas ($\Delta x, \Delta y, \Delta z$), axis-angle rotation deltas ($\Delta r_x, \Delta r_y, \Delta r_z$), and gripper state. We use a batch size of 24 for LIBERO and 16 for CALVIN, a learning rate of $5 \times 10^{-5}$, $K_L = 2$ executor iterations per cycle, and $K_H = 1$ reasoning cycle per segment. The head hidden dimension is $D = 896$ and the chunk length is $N_{\text{act}} = 8$, and the backbone carries $N_q = 64$ action-query tokens. We use $N_{\text{plan}} = 32$ high-level planning tokens and $N_{\text{viz}}=512$ visual tokens. The primary visual features are extracted from the penultimate backbone layer.
During training, we unroll all $S_{\text{total}}=8$ reasoning segments and apply deep supervision to the intermediate action predictions. At inference, adaptive halting is configured with a minimum of $S_{\min}=2$ segments, a maximum of $S_{\max}=8$ segments, a prediction-stability threshold of $\tau=0.01$, and a patience of $P=1$.

\subsection{Simulation Results}
\label{sec:sim_results}

\begin{table*}[t]
    \centering
    \small
    \renewcommand{\arraystretch}{1.2}
    \setlength{\tabcolsep}{5pt}

    \caption{LIBERO success rates (\%) by suite.
    The number after each model name is the pretrained VLM size in billions of parameters.
    Head (M) denotes the action-head size in millions of parameters.
    Head size governs the memory footprint of the reasoning module on resource-constrained hardware.
    -- indicates that a separately identifiable action head is not available. 
    \textbf{Bold}: best; \underline{\textit{italics}}: second best. VLA-Adapter results are our reproduction of the released checkpoint.}
    \label{tab:libero_comparison}

    \begin{tabular}{l l c | c c c c | c}
        \toprule
        \textbf{Method} & \textbf{Model} &
        \textbf{Head (M)} &
        \textbf{Spat.} & \textbf{Obj.} &
        \textbf{Goal} & \textbf{Long} & \textbf{Avg.} \\
        \midrule

        \multirow{6}{*}{\textbf{E2E VLAs}}
            & SmolVLA-0.5B~\cite{shukor2025smolvla}
            & 98.2 & 90.0 & 96.0 & 92.0 & 71.0 & 87.3 \\

            & OpenVLA-7B~\cite{kim2024openvla}
            & -- & 84.7 & 88.4 & 79.2 & 53.7 & 76.5 \\

            & WorldVLA-7B~\cite{cen2025worldvla}
            & -- & 87.6 & 96.2 & 83.4 & 60.0 & 81.8 \\

            & $\pi_0$-FAST-3B~\cite{pertsch2025fast}
            & -- & 96.4 & 96.8 & 88.6 & 60.2 & 85.5 \\

            & VLA-Adapter-0.5B~\cite{wang2026vlaadapter}
            & 97 & \underline{\textit{97.2}} & \underline{\textit{99.2}} & 96.0 & \textbf{94.0} & \underline{\textit{96.6}} \\

            & SpatialVLA-4B~\cite{qu2025spatialvla}
            & -- & 88.2 & 89.9 & 78.6 & 55.5 & 78.1 \\

        \cmidrule{1-8}

        \multirow{6}{*}{\textbf{CoT Reasoning}}
            & CoT-VLA-7B~\cite{zhao2025cot}
            & -- & 87.5 & 91.6 & 87.6 & 69.0 & 81.1 \\

            & FlowVLA-8.5B~\cite{zhong2026flowvla}
            & -- & 93.2 & 95.0 & 91.6 & 72.6 & 88.1 \\

            & ThinkAct-7B~\cite{huang2025thinkact}
            & 432 & 88.3 & 91.4 & 87.1 & 70.9 & 84.4 \\

            & Fast-ThinkAct-3B~\cite{huang2026fast}
            & $\sim$1200 & 92.0 & 97.2 & 90.2 & 79.4 & 89.7 \\

            & TraceVLA-7B~\cite{zheng2025tracevla}
            & -- & 84.6 & 85.2 & 75.1 & 54.1 & 74.8 \\

            & MolmoAct-7B~\cite{lee2026molmoact}
            & -- & 87.0 & 95.4 & 87.6 & 77.2 & 86.6 \\

        \cmidrule{1-8}

        \multirow{3}{*}{\textbf{Iterative Refinement}}
            & RD-VLA-0.5B (Adapt.)~\cite{tur2026rdvla}
            & 72.3 & 88.6 & 98.8 & \underline{\textit{96.8}} & 85.8 & 92.5 \\

            & LoopOFT$\alpha$-1.2B~\cite{shen2026loopvla}
            & -- & 94.0 & \textbf{99.6} & \underline{\textit{96.8}} & 91.0 & 95.3 \\

            \cmidrule{2-8}

            & \textbf{HR-AH-0.5B (Ours)}
            & \textbf{22.5}
            & \textbf{98.0} & \textbf{99.6} & \textbf{97.4} & \underline{\textit{93.0}} & \textbf{97.0} \\
        \bottomrule
    \end{tabular}
\end{table*}

\paragraph{LIBERO.}
\label{sec:libero}
Table~\ref{tab:libero_comparison} compares our method against state-of-the-art end-to-end (E2E) VLAs, Chain-of-Thought (CoT) reasoning models, and iterative refinement baselines on the LIBERO benchmarks. HR-AH achieves the highest average success rate (97.0\%) across the four suites, leading on Spatial (98.0\%), tying on Object (99.6\%), and taking the top score on Goal (97.4\%), while placing second on Long (93.0\%) only behind VLA-Adapter. The HR-AH action head contains only 22.5M parameters, roughly $4\times$ smaller than VLA-Adapter's 97M head and $3\times$ smaller than RD-VLA's 72.3M head. HR-AH also achieves the lowest inference latency among the methods compared, at 49.63\,ms, as shown in \cref{fig:latency_success_rate}.

\paragraph{CALVIN.}
\label{sec:calvin}
We further evaluate on CALVIN ABC$\rightarrow$D, which tests generalization to unseen environments. Table~\ref{tab:calvin_comparison} reports success rates for chains of up to five tasks and the average chain length. HR-AH achieves an average chain length of 4.40, matching VLA-Adapter while using a $4\times$ smaller head, and substantially outperforms RD-VLA (3.39) and OpenVLA (3.27). Compared to RD-VLA, the nearest recurrent head on the same backbone, the gains are especially pronounced on longer chains, with HR-AH completing 5-task chains in 74.0\% of rollouts versus 45.3\%.

This gain reflects the benefit of the high-level planning state $z_H$ within each action chunk. While the executor refines the motor command over multiple fast steps, $z_H$ integrates the task-relevant visual context once per slow-rate update and holds it stable across the remaining fast steps. The executor can therefore focus on precise action refinement without reconstructing the scene at every step, an advantage over the single-state design that compounds over the longer rollouts required by multi-task chains.

\begin{table}[t]
    \centering
    \renewcommand{\arraystretch}{1.0}
    
    \captionsetup{width=\linewidth} 
    \caption{CALVIN ABC$\rightarrow$D results. We report the percentage of successful chains of length 1 to 5, and the average chain length. 
    The number after each model name is the pretrained VLM size in billions of parameters. \textbf{Bold}: best; \underline{\textit{italics}}: second best. VLA-Adapter results are our reproduction of the released checkpoint.}
    \label{tab:calvin_comparison}

    \resizebox{\columnwidth}{!}{%
    \small
    \begin{tabular}{@{} l c c c c c c @{}}
        \toprule
        \multirow{2}{*}{\textbf{Method}} & \multicolumn{5}{c}{\textbf{Tasks completed in a row $\uparrow$}} & \multirow{2}{*}{\textbf{Avg. len $\uparrow$}} \\
        \cmidrule(lr){2-6}
         & \textbf{1} & \textbf{2} & \textbf{3} & \textbf{4} & \textbf{5} &  \\
        \midrule

        OpenVLA-7B~\cite{kim2024openvla} & 91.3 & 77.8 & 62.0 & 52.1 & 43.5 & 3.27 \\
        VLAS-7B~\cite{zhao2025vlas} & 87.2 & 64.2 & 40.9 & 28.1 & 19.6 & 2.40 \\
        LCB-7B~\cite{shentu2024llms} & 73.6 & 50.2 & 28.5 & 16.0 & 9.9 & 1.78 \\
        VLA-Adapter-0.5B~\cite{wang2026vlaadapter} & \underline{\textit{97.2}} & \underline{\textit{93.8}} & \textbf{89.1} & \underline{\textit{83.2}} & \textbf{76.3} & \textbf{4.40} \\
        \midrule

        DeeR-3B~\cite{yue2024deervla} & 86.2 & 70.1 & 51.8 & 41.5 & 30.4 & 2.82 \\
        RoboFlamingo-3B~\cite{li2024vision} & 82.4 & 61.9 & 46.6 & 33.1 & 23.5 & 2.48 \\
        GR-1-3.1B~\cite{wu2024unleashing} & 85.4 & 71.2 & 59.6 & 49.7 & 40.1 & 3.06 \\
        SuSIE-1.3B~\cite{black2024zero} & 87.0 & 69.0 & 49.0 & 38.0 & 26.0 & 2.69 \\
        \midrule

        RT-1~\cite{brohan2022rt} & 53.3 & 22.2 & 9.4 & 3.8 & 1.3 & 0.90 \\
        RD-VLA-0.5B~\cite{tur2026rdvla} & 91.4 & 79.5 & \underline{\textit{67.9}} & 54.9 & 45.3 & \underline{\textit{3.39}} \\
        \textbf{HR-AH-0.5B (Ours)} & \textbf{99.3} & \textbf{94.6} & \textbf{89.1} & \textbf{83.4} & \underline{\textit{74.0}} & \textbf{4.40} \\
        
        \bottomrule
    \end{tabular}%
    }
\end{table}

\subsection{Real-World Results}
\label{sec:real_world}
\begin{figure}[t]
    \centering
    \includegraphics[width=0.75\columnwidth, trim=10cm 1.5cm 2cm 1.5cm, clip]{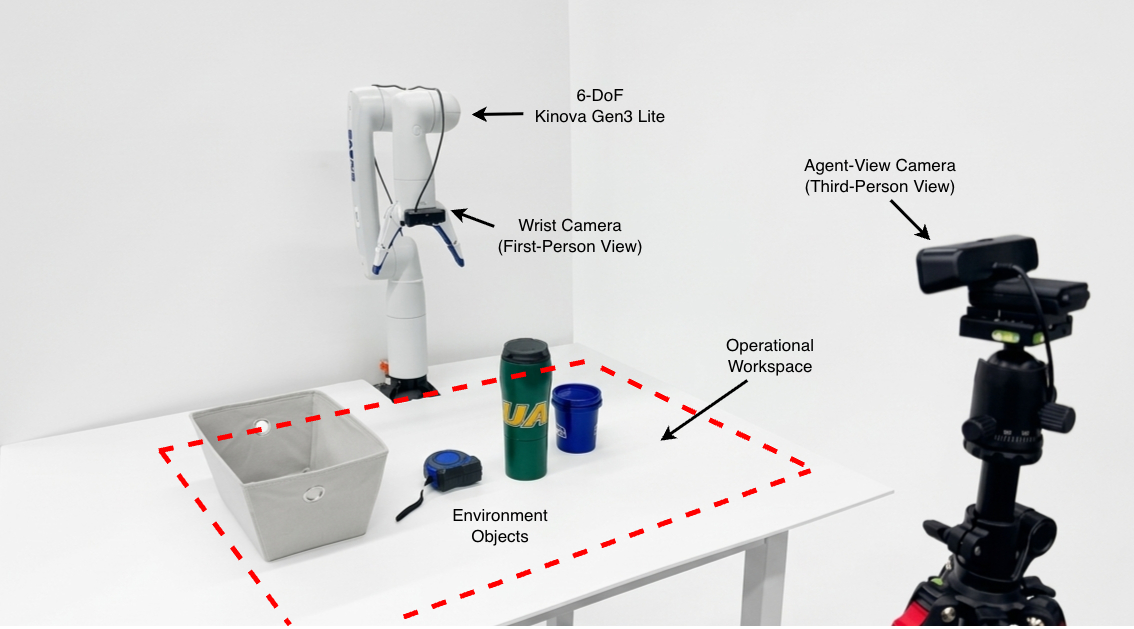}     
    \caption{Real-world experimental setup using the Kinova Gen3 Lite robot arm.}
    \label{fig:real_world}
\end{figure}

\begin{figure*}[t]
    \centering
    \setlength{\tabcolsep}{1pt}
    \setlength{\rowlabw}{14pt}
    \setlength{\imgw}{\dimexpr(0.95\textwidth - \rowlabw - 40\tabcolsep)/6\relax}

    \begin{tabular}{@{}c c c c c c c@{}}
        \raisebox{0.38\imgw}{\rotatebox{90}{\footnotesize\bfseries RD-VLA}} &
        \includegraphics[width=\imgw]{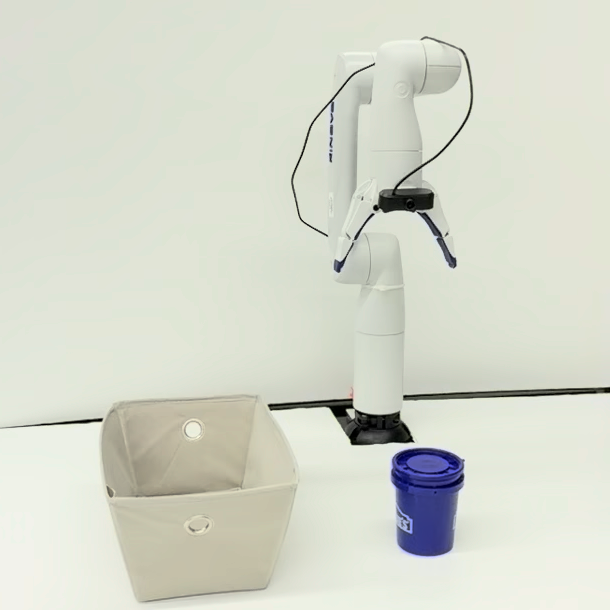} &
        \includegraphics[width=\imgw]{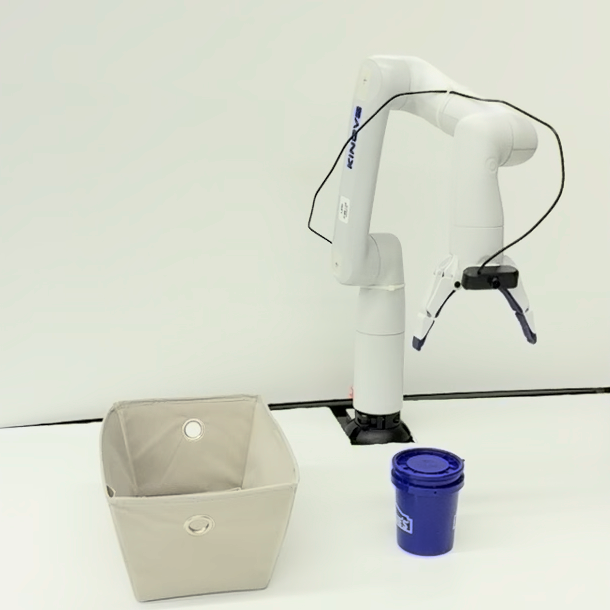} &
        \includegraphics[width=\imgw]{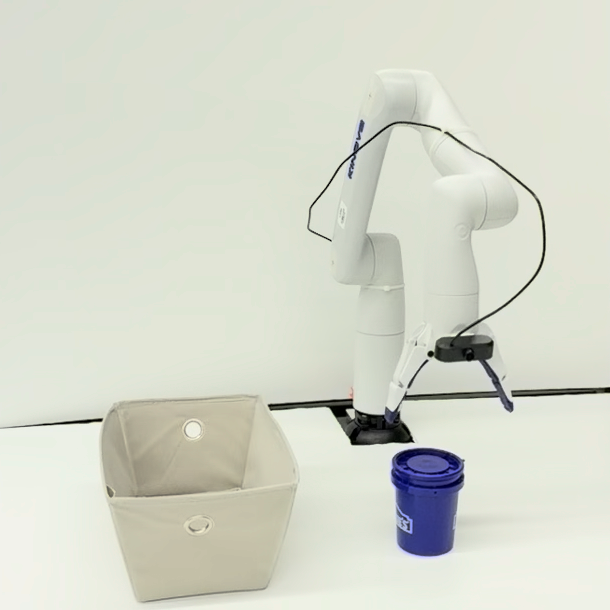} &
        \includegraphics[width=\imgw]{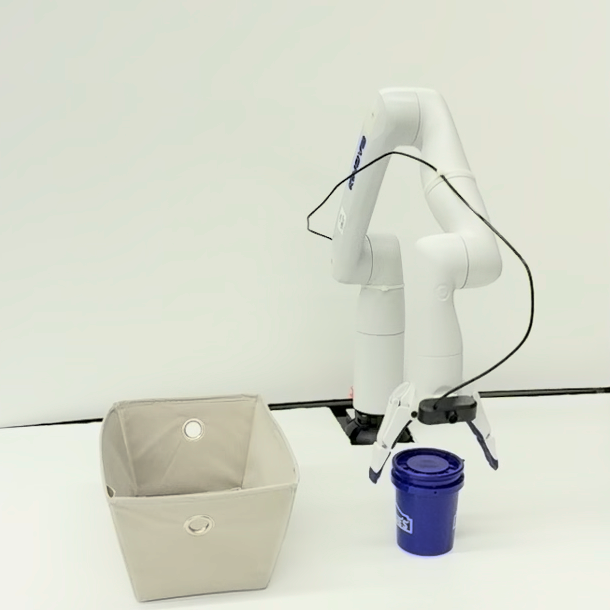} &
        \includegraphics[width=\imgw]{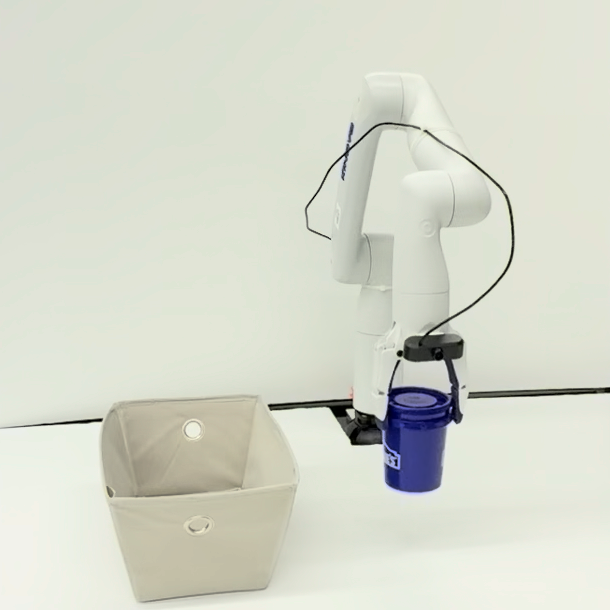} &
        \includegraphics[width=\imgw]{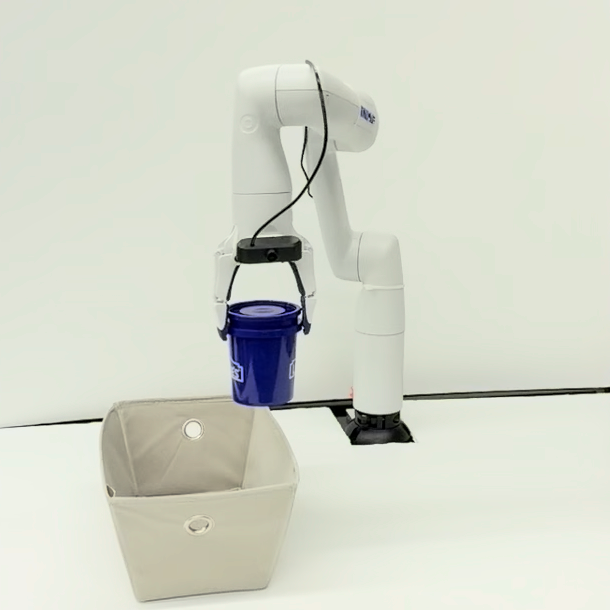}
        \\[2pt]
        \raisebox{0.38\imgw}{\rotatebox{90}{\footnotesize\bfseries Ours}} &
        \includegraphics[width=\imgw]{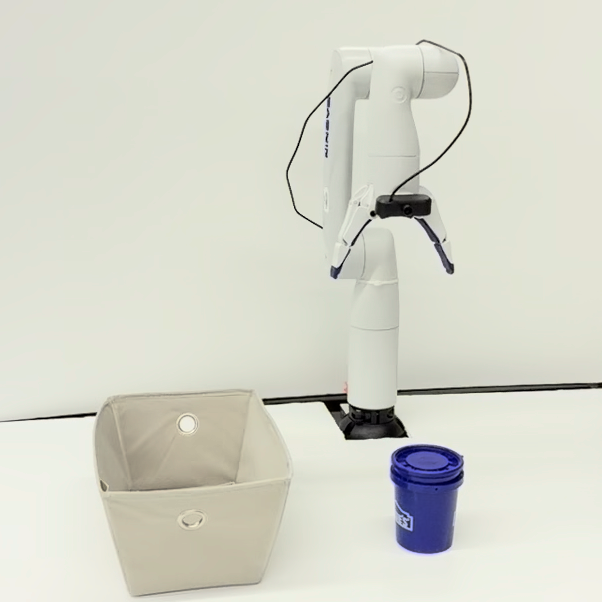} &
        \includegraphics[width=\imgw]{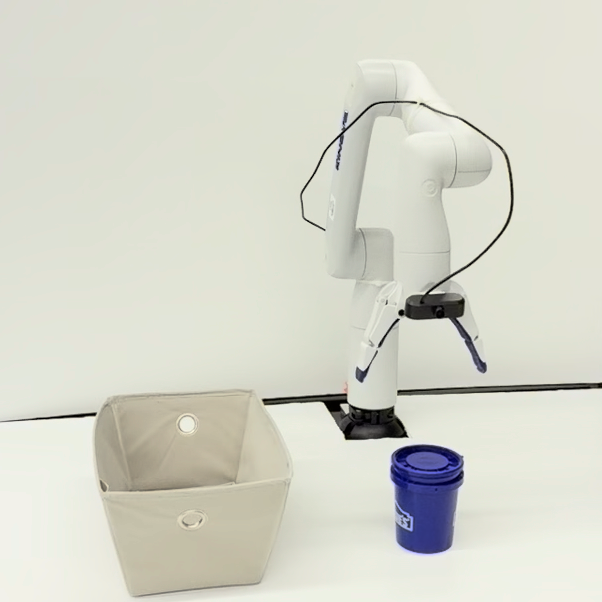} &
        \includegraphics[width=\imgw]{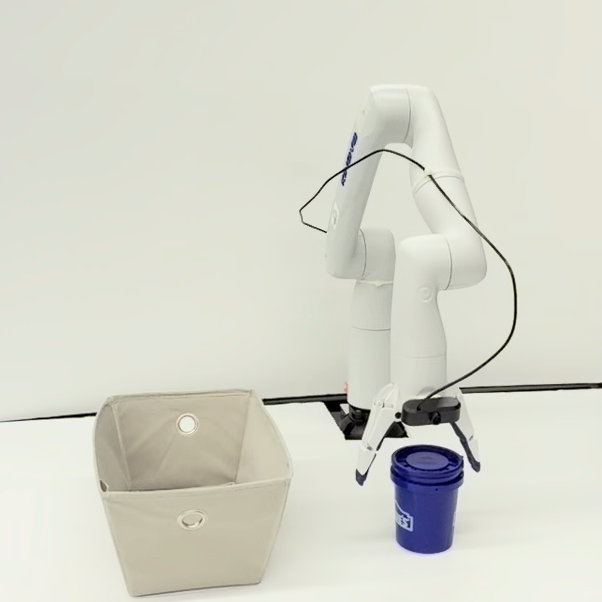} &
        \includegraphics[width=\imgw]{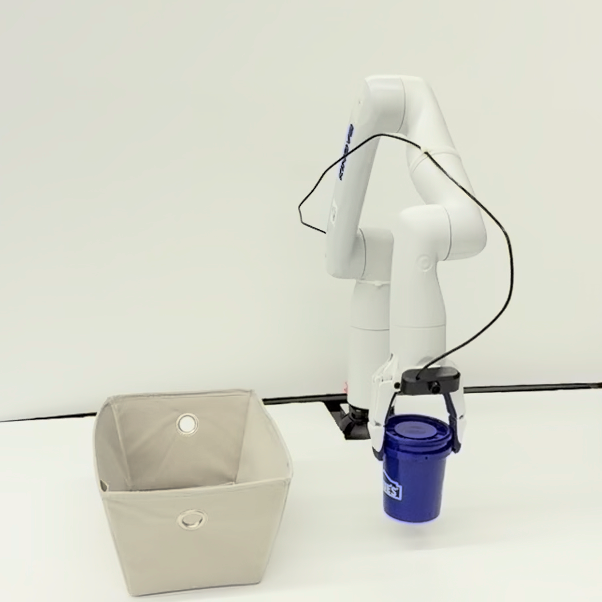} &
        \includegraphics[width=\imgw]{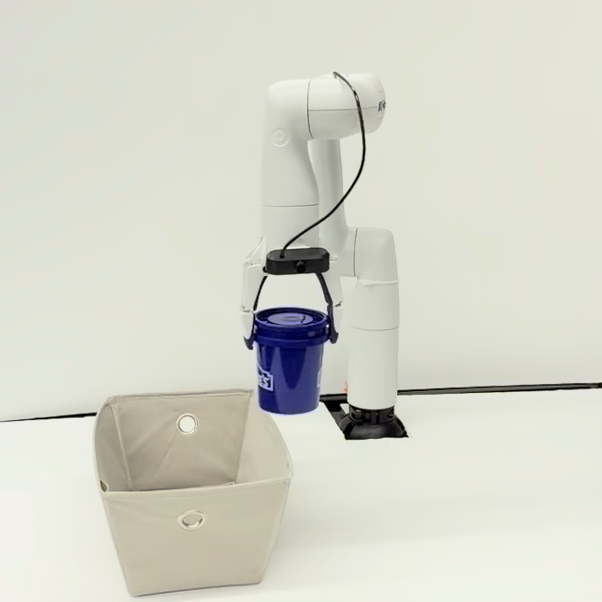} &
        \includegraphics[width=\imgw]{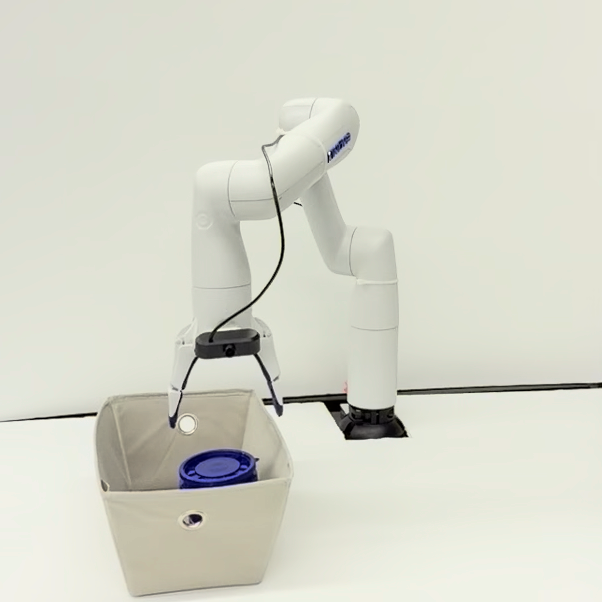}
        \\[2pt]
        & {\small $t=0$} & {\small $t=4$} & {\small $t=7$} &
          {\small $t=11$} & {\small $t=14$} & {\small $t=18$}
    \end{tabular}

    \caption{Real-world rollout comparison between RD-VLA (top) and HR-AH (bottom) on the \emph{Put} task. Both methods succeed in this example, isolating the difference in execution speed.}
    \label{fig:rw_rollout_comparison}
\end{figure*}

To evaluate physical generalizability, we collect demonstrations on a 6-DoF Kinova Gen3 Lite robotic arm across four tasks of increasing complexity: \textit{Put}, \textit{Move}, \textit{Stack}, and \textit{Long}, testing basic manipulation, spatial reasoning, and sustained behavior over long sequences. We collect 20 demonstrations per task for Put, Move, and Stack, and 30 for Long. Each policy is evaluated over 20 rollouts per task.
Figure~\ref{fig:real_world} shows the setup and Table~\ref{tab:real_world} reports success rates for HR-AH and RD-VLA, the two recurrent-head methods on the same backbone.
Beyond higher accuracy, HR-AH also executes faster. Figure~\ref{fig:rw_rollout_comparison} shows that RD-VLA requires more time to complete the same task, consistent with the latency gap observed in simulation (\cref{fig:latency_success_rate}).

\begin{table}[t]
    \centering
    \small
    \renewcommand{\arraystretch}{1.1}
    \setlength{\tabcolsep}{4pt}
    \caption{Real-world success rates (\%) on the Kinova Gen3 Lite. We evaluate four manipulation tasks of increasing complexity.}
    \label{tab:real_world}
    \begin{tabular}{l c c c c | c}
        \toprule
        \textbf{Method} & \textbf{Put} & \textbf{Move} & \textbf{Stack} & \textbf{Long} & \textbf{Avg.} \\
        \midrule
        RD-VLA~\cite{tur2026rdvla} & 85.0 & 75.0 & 60.0 & 45.0 & 66.3 \\
        \textbf{HR-AH (Ours)} & \textbf{95.0} & \textbf{90.0} & \textbf{75.0} & \textbf{55.0} & \textbf{78.8} \\
        \bottomrule
    \end{tabular}
\end{table}

\subsection{Ablation Studies}
\label{sec:ablations}

\begin{figure}[t]
    \centering
    \includegraphics[
        width=\columnwidth
    ]{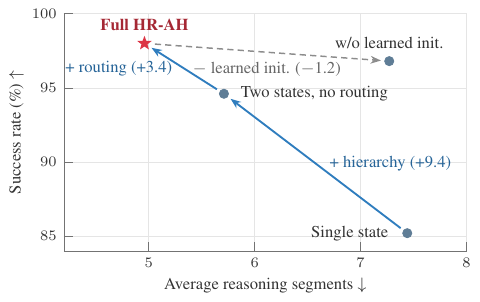}

    \caption{
        Architectural ablation on LIBERO-Spatial.
        Each point plots success rate against average reasoning segments (upper left is best). Solid arrows follow the additive build-up from a parameter-matched single-state baseline to the full model, and the dashed arrow shows the effect of removing learned initialization. Hierarchical recurrence raises success from 85.2\% to 94.6\% while reducing average segments from 7.44 to 5.71. Modality-specific routing further raises success to 98.0\% and reduces segments to 4.96. Removing learned initializations drops success to 96.8\% and raises average segments to 7.27.
    }

    \label{fig:ablation_arch}
\end{figure}

\cref{fig:ablation_arch} isolates the contributions of hierarchy, modality routing, and learned initialization on LIBERO-Spatial. Compared with a parameter-matched single-state baseline receiving full context, introducing a second recurrent state with identical inputs improves success by 9.4 percentage points while reducing the average number of reasoning segments. Modality-specific routing contributes a further 3.4-point gain and reduces inference depth to 4.96 segments. Removing learned state initializations lowers success by 1.2 points and increases the average segments to 7.27. These results indicate that hierarchical state separation provides the largest gain, with routing providing a substantial further gain and learned initialization offering an additional improvement. \cref{fig:qualitative_comparison} reflects the effect qualitatively: HR-AH concentrates attention on task-relevant objects over successive reasoning steps, whereas the single-state baseline attends more diffusely.

\subsection{Inference Efficiency}
\begin{figure}[t]
    \centering
    \begin{tikzpicture}
    \begin{axis}[
        width=\columnwidth, height=6.6cm,
        xmode=log, xmin=30, xmax=8000, ymin=85, ymax=100,
        xtick={100,1000,8000}, xticklabels={100,1000,8000},
        ytick={85,90,95,100},
        xlabel={Latency (ms, log scale)}, ylabel={Success rate (\%)},
        label style={font=\footnotesize}, tick label style={font=\scriptsize},
        axis x line*=bottom, axis y line*=left,
        axis line style={black!55}, tick style={black!55},
        grid=major, major grid style={black!10, line width=0.3pt},
        xlabel style={yshift=2pt}, ylabel style={yshift=-3pt},
        clip=false,
        nodes={font=\footnotesize, text=black!80},
    ]
        \addplot[only marks, mark=*, mark size=2.2pt, color=plotblue] coordinates {(52.63,97.2)};
        \addplot[only marks, mark=triangle*, mark size=2.8pt, color=plotblue!55!black!70] coordinates {(810,92.0) (6800,87.0) (7600,88.3)};
        \addplot[only marks, mark=square*, mark size=2pt, color=plotblue!45] coordinates {(85,88.6)};
        \addplot[only marks, mark=fstar, mark size=2.6pt, color=plotred] coordinates {(49.63,98.0)};

        \node[anchor=west, text=plotred!75!black, font=\footnotesize\bfseries] at (axis cs:57,98.5) {HR-AH (ours, 22.5M)};
        \node[anchor=west] at (axis cs:61,96.9) {VLA-Adapter (97M)};
        \node[anchor=north] at (axis cs:118,88.25) {RD-VLA (adaptive, 72.3M)};
        \node[anchor=south east] at (axis cs:760,92.3) {Fast-ThinkAct-3B ($\sim$1200M)};
        \node[anchor=east] at (axis cs:5800,86.6) {MolmoAct-7B};
        \node[anchor=east] at (axis cs:6500,88.7) {ThinkAct-7B (432M)};

        \draw[-{Stealth[length=4pt]}, black!40, line width=0.7pt] (axis cs:3500,92.5) -- (axis cs:1400,95.8)
            node[midway, sloped, below, black!50, font=\scriptsize\itshape] {better};
    \end{axis}
    \end{tikzpicture}

    \caption{Success rate versus latency on LIBERO-Spatial. HR-AH (star) achieves 98.0\% at 49.63\,ms, combining the highest accuracy with the lowest latency among compared methods. Circle: end-to-end, triangle: CoT reasoning, square: iterative refinement. Numbers in parentheses give the action-head size. 
    }
    \label{fig:latency_success_rate}
\end{figure}
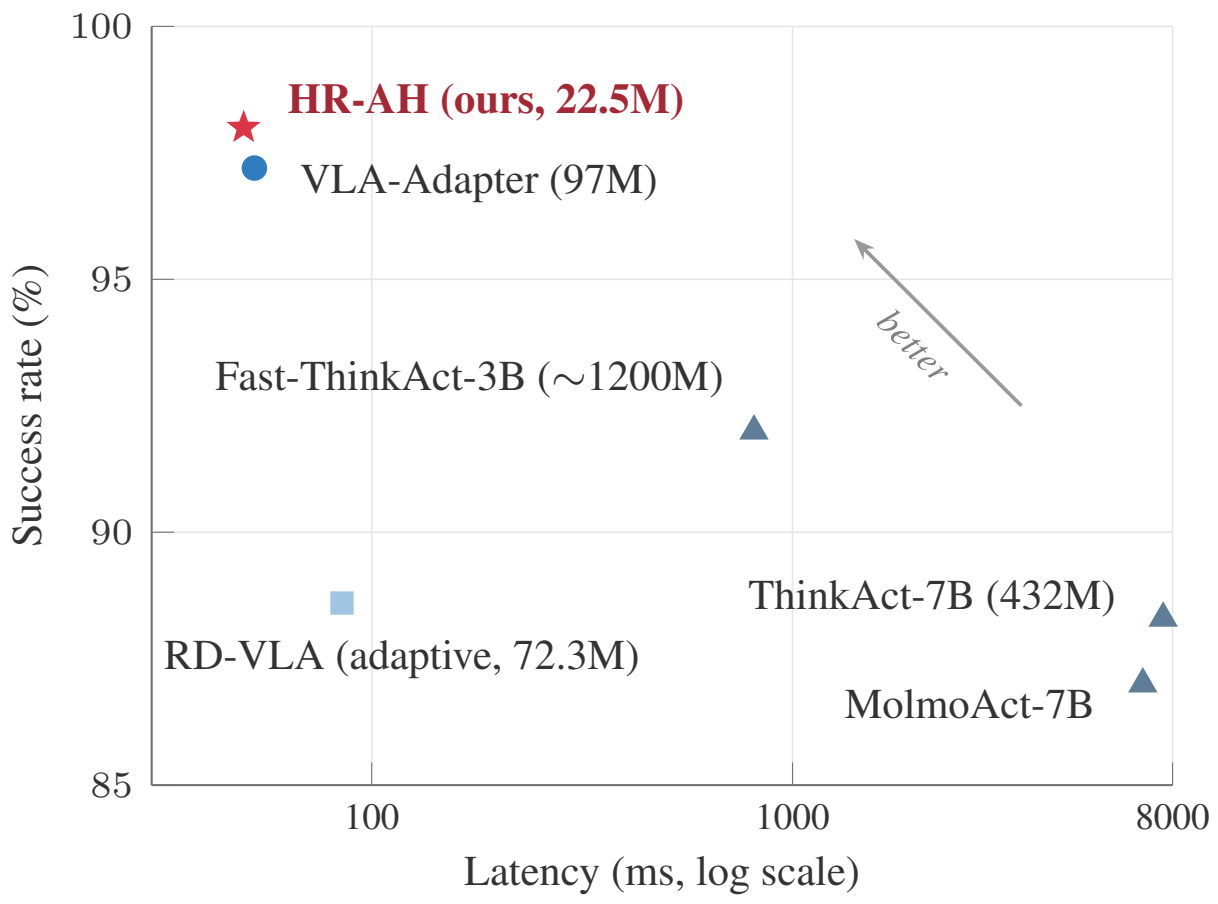

Table~\ref{tab:ablation_efficiency} analyzes the trade-off between computation and accuracy. Accuracy is not monotonic in the number of segments: Fixed-4 (96.4\%) falls below Fixed-2 (97.0\%), while Fixed-8 reaches the peak (98.2\%), suggesting that no single fixed budget suits all tasks. Our adaptive halting strategy ($\tau=0.01$) reaches 98.0\% using only 4.96 segments on average, a 38\% reduction over Fixed-8 at a cost of only 0.2 points. By letting each observation choose its own depth, adaptive halting avoids committing to a fixed budget that may be too large for easy actions or too small for hard ones. \cref{fig:latency_success_rate} plots success rate against per-chunk latency on LIBERO-Spatial for representative end-to-end, CoT reasoning, and iterative refinement methods.

\begin{table}[t]
\centering
\caption{Fixed segment budgets versus adaptive halting for HR-AH on LIBERO-Spatial.}
\label{tab:ablation_efficiency}
\resizebox{\columnwidth}{!}{%
\begin{tabular}{lcc}
\toprule
\textbf{Inference Strategy} &
\textbf{Success Rate (\%)} &
\textbf{Avg. Segments}\\
\midrule
Fixed-2 & 97.0 & \textbf{2.00} \\
Fixed-4 & 96.4 & 4.00 \\
Fixed-6 & 97.4 & 6.00 \\
Fixed-8 & \textbf{98.2} & 8.00 \\
Adaptive ($\tau=0.05$) & 97.6 & 2.89\\
\textbf{Adaptive ($\tau=0.01$)} & 98.0 & 4.96 \\
\bottomrule
\end{tabular}%
}
\end{table}

\section{Conclusion}
We introduced the Hierarchical Reasoning Action Head (HR-AH), a lightweight recurrent module that brings structured deliberation to tiny VLA models. By decoupling the latent state into a slow planner and a fast executor, HR-AH bridges the gap between compact model size and the need for iterative reasoning.
Among the methods compared, HR-AH achieves the highest accuracy on LIBERO and matches the best chain length on CALVIN with an action head $3$--$4.3\times$ smaller than prior recurrent and feed-forward heads on the same backbone, and the gains transfer to a physical robot. Ablations show that the improvement stems from both the hierarchical state separation and the modality-specific routing, which confines costly vision attention to the slow planner. Together, these results suggest that structured deliberation in a compact head is sufficient for strong performance on a tiny backbone, opening a path toward capable VLAs deployable on resource-constrained robots.

{
    \small
    \bibliographystyle{ieeenat}
    \bibliography{main}
}

\end{document}